\pdfoutput=1

\documentclass[11pt]{article}
\usepackage{booktabs} 
\usepackage{acl}
\usepackage{multirow} 
\usepackage{graphicx}
\usepackage[most]{tcolorbox}
\usepackage{times}
\usepackage{latexsym}
\usepackage{tcolorbox}
\usepackage{booktabs} 
\usepackage{capt-of}
\usepackage{placeins}
\usepackage{cuted}
\usepackage[T1]{fontenc}
\usepackage{multicol}
\usepackage[utf8]{inputenc}

\usepackage{microtype}

\title{PPTC-R benchmark: Towards Evaluating the Robustness of Large Language Models for PowerPoint Task Completion}

\author{
Zekai Zhang$^{1}$\footnotemark[1],~~Yiduo Guo$^{1}$\footnotemark[1],~~Yaobo Liang$^{2}$,~~Dongyan Zhao$^{1}$,~~Nan Duan$^{2}$\\ 
$^1$Peking University\\
$^2$Microsoft Research Asia\\
\texttt{\{justinzzk,yiduo\}@stu.pku.edu.cn,\{yaobo.liang,nanduan\}@microsoft.com}\\\texttt{zhaody@pku.edu.cn}\\
}

\begin{document}
\maketitle
\renewcommand{\thefootnote}{\fnsymbol{footnote}}
\footnotetext[1]{Equal contribution}
\begin{abstract}
The growing dependence on Large Language Models (LLMs) for finishing user instructions necessitates a comprehensive understanding of their robustness to complex task completion in real-world situations. To address this critical need, we propose the PowerPoint Task Completion-Robustness (PPTC-R) benchmark to measure LLMs’ robustness to the user PPT task instruction and software version (Powerpoint). Specifically, we construct adversarial user instructions by attacking user instructions at sentence, semantic, and multi-language levels. 
To assess the robustness of Language Models to software versions, we vary the number of provided APIs to simulate both the newest version and earlier version settings. Subsequently, we test 3 closed-source and 4 open-source LLMs using a benchmark that incorporates these robustness settings, aiming to evaluate how deviations impact LLMs' API calls for task completion. We find that GPT-4 exhibits the highest performance and strong robustness in our benchmark, particularly in the version update and the multilingual settings. However, we find that all LLMs lose their robustness when confronted with multiple challenges (e.g., multi-turn) simultaneously, leading to significant performance drops. We further analyze the robustness behavior and error reasons of LLMs in our benchmark, which provide valuable insights for researchers to understand the LLM's robustness in task completion and develop more robust LLMs and agents. We release the code and data at \url{https://github.com/ZekaiGalaxy/PPTCR}. 
\end{abstract}
\section{Introduction}
\begin{figure}[ht]
  \centering
  \includegraphics[width=0.5\textwidth]{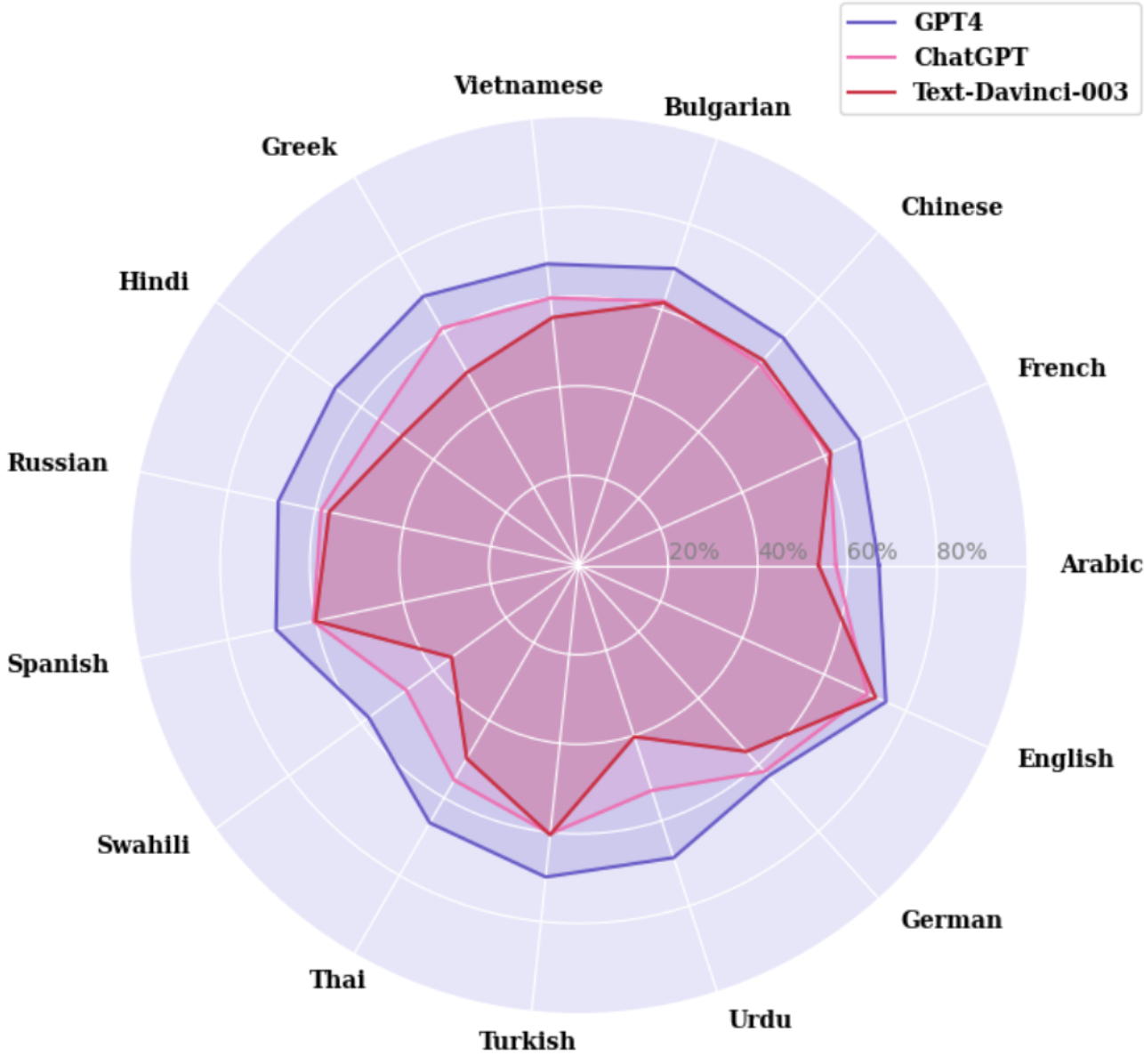}
  \caption{We illustrate the turn-base multilingual results of closed-source LLMs.}
  \label{fig:multilingual}
\vspace{-10pt}
\end{figure}
\begin{figure*}[ht]
  \centering
  \includegraphics[width=1\textwidth]{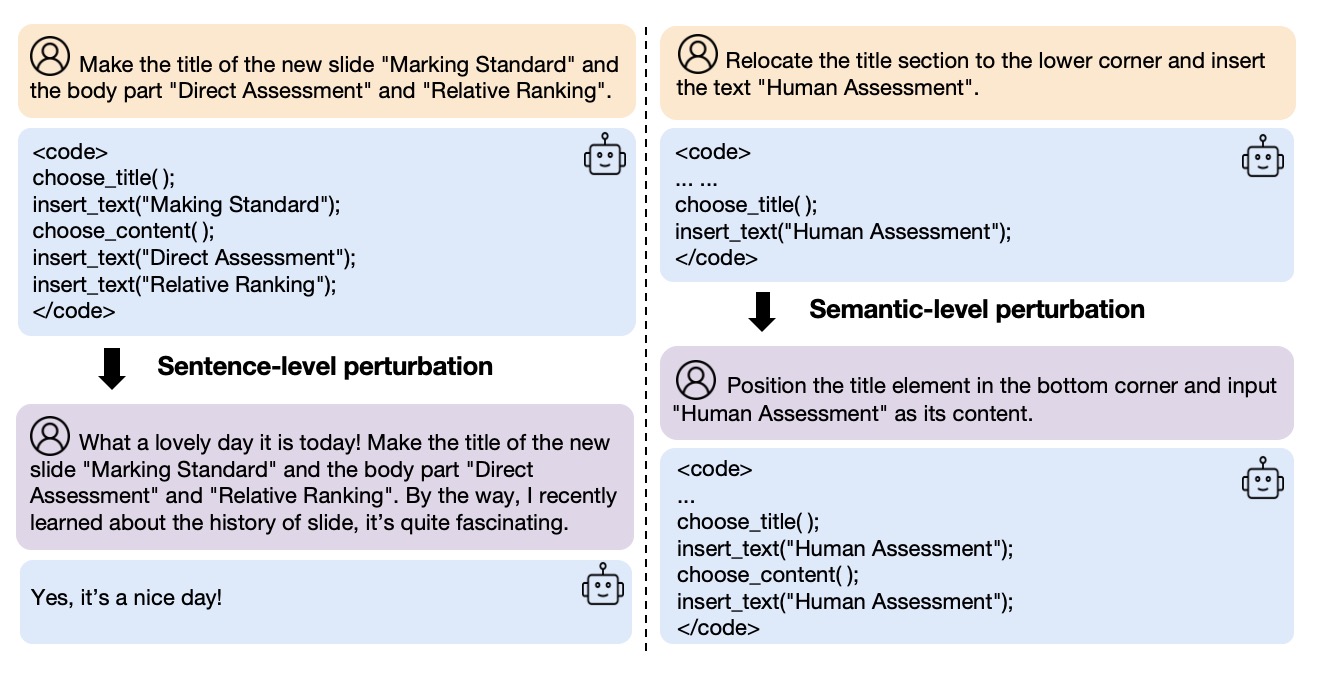}
  \caption{We illustrate two examples for constructing our robustness benchmark. The perturbations correctly distract the LLM from completing the user instruction (the left) and mislead the LLM into generating the wrong API sequence (the right), which underscores the importance of evaluating and analyzing LLMs' task completion robustness.}
  \label{fig:demonstration}
\vspace{-10pt}
\end{figure*}
 Large Language Models (e.g. GPT-4~\cite{openai2023gpt4}) show strong strong performance on various basic natural language tasks and human examinations~\cite{qin2023chatgpt,jiao2023chatgpt,zhong2023agieval,wang2023chatgpt,liang2023taskmatrix}, and arises the hope to help humans to complete tasks in complex environments, such as purchasing items in WebShop~\cite{yao2022webshop}, creating and editing PPT slide in PPTC~\cite{Guo2023PPTCBE}, and nagativing computer in MiniWob++~\cite{liu2018reinforcement}. Also, recent works~\cite{zhu2023promptbench,liu2023robustness,wang2023robustness} such as PromptBench~\cite{zhu2023promptbench} study the LLM's robustness to task prompts for basic natural language tasks. However, there remains an absence of a benchmark for evaluating the LLM's robustness in complex task completion, which is a key factor for the LLM's task completion performance in real-world user scenarios. To address this need, We introduce
\textbf{P}ower\textbf{P}oint \textbf{T}ask \textbf{C}ompletion-\textbf{R}obustness (PPTC-R), a benchmark for measuring and analyzing LLMs' robustness to the user instruction and software version in PowerPoint task completion setting. Our benchmark has two distinct features: (1) Previous robustness evaluations are based on traditional natural language tasks, where the model only needs to generate options or text strings. In contrast, we focus on evaluating how adversarial perturbation influences LLMs' API calls for complex PPT task completion. (2) Previous studies construct their benchmarks mainly by attacking the task prompt or input in text. We consider how the shift of the software version influences the LLM's performance, which is a new perspective.

To measure LLMs' robustness to the user instruction, we need to construct the adversarial user instruction. We 
consider instruction perturbations including 1) (language-level) translating the original English instructions into 14 non-English languages (See Figure~\ref{fig:multilingual}), 2) (sentence-level) adding GPT-4 generated chitchat sentences into the original instructions as noisy sentences, and 3) (semantic-level) prompting GPT-4 to express
original instructions with the same semantic meaning in 4 different ways (See examples in Figure~\ref{fig:demonstration}). These perturbations commonly occur to normal users or developers in their daily use of PPT (with agent). Our LLM-based sentence and semantic perturbation approaches can quickly obtain lots of new high-quality adversarial data without being seen before. On the other hand, the software versions can affect the PPT task completion process by providing different numbers of functions (APIs) for the user. Thus we test the LLM's robustness to the software version by adjusting the number of provided APIs: (1) introducing many new APIs into the existing API list to simulate the version update situation where new APIs may impact the LLM's API selection and (2) removing many APIs in the existing API list to simulate the situation in which APIs in the current software version may not be capable of completely addressing some user instruction and the LLM needs to seek for assistance. We conduct these perturbations (~3 for user instructions and 2 for APIs) on the original PPTC benchmark separately to construct our PPTC-R benchmarks (a total of 5 settings). 

We test 3 closed-source LLMs (e.g., GPT-4 and ChatGPT) and 4 representative open-source LLMs (e.g., LLaMa-2 and WizardLM) in our benchmarks. We find that GPT-4 achieves the highest performance and strong robustness in our 5 settings (see Sec.~\ref{sec.results}). For example, GPT-4 can maintain its high turn-based performance with the introduction of 97 new APIs.  In contrast, other LLMs experience a larger performance drop (e.g., ChatGPT) or maintain their performance at a pretty low level (e.g., LLaMa-2). We also find a unique robustness degradation phenomenon for all LLMs: The LLMs' robustness decreases obviously when we increase the difficulty of the same task or move to a more complex environment. We further analyze and find three main error reasons for LLMs: being distracted by chitchat (See the bottom of Figure~\ref{fig:demonstration}), calling unavailable APIs, and misunderstanding instructions with new expressions (refer to Sec.~\ref{sec.error}). We also investigate the LLM's behavior with different numbers of new APIs (See Sec~\ref{sec.api_number}). 

In summary, the contributions of our paper are:

(1) We propose the PowerPoint Task Completion Robustness benchmark, which is the first one to measure LLM's task completion robustness for calling APIs to complete user instructions. Furthermore, our LLM-based perturbation approaches can be easily deployed to generate adversarial data for future datasets.

(2) We test 7 LLMs in our benchmark and find that GPT-4 achieves the best performance with strong robustness. However, all LLMs' robustness degrades when we increase the task difficulty, showing the challenge of our benchmark.

(3) We further analyze the error reasons and robustness behavior of LLMs in our benchmark, which provides valuable insights for researchers to understand LLMs' robustness in task completion settings and to design more robust agent.

\section{Related Works}

\textbf{Large Language Models (LLMs)} such as GPT-4~\cite{bubeck2023sparks,openai2023gpt4}, and PaLM-2~\cite{anil2023palm} exhibit excellent performance for various traditional natural language tasks~\cite{kim2023language,jiao2023chatgpt,zhong2023agieval,wang2023chatgpt} and can do complex logic reasoning~\cite{feng2023language,liu2023evaluating}, pass human-level examination~\cite{zhong2023agieval,gilson2023does,katz2023gpt}, and write code~\cite{li2022competition,liu2023your} after instruction fine-tuning. Open-source LLMs like LLaMa-2~\cite{touvron2023llama}, Mistral 7b~\cite{jiang2023mistral}, and Baichuan-2~\cite{yang2023baichuan} and their fine-tuned versions also show promising performance on public benchmarks. Recent survey~\cite{chen2023chatgpt} finds that they usually still have a performance gap when compared to their closed-source counterparts like GPT-4.  

\textbf{Task completion benchmarks for LLM-based Agents}. LLMs and multi-modal models (e.g., GPT-4Vision) raise the hope of designing LLM-based agents to help humans finish complex tasks in complex environments.  To test agents,
Saycan~\cite{brohan2023can}, Behavior~\cite{srivastava2022behavior,li2023behavior} and VirtualHome~\cite{puig2018virtualhome} benchmarks ask the agent to negative a series of physical actions to finish the user instruction in simulated physical environments. WebShop~\cite{yao2022webshop}, AgentBench~\cite{liu2023agentbench} and  Android in the wild~\cite{rawles2023android} require the agent conduct actions (e.g., click and search) in website environment to meet the user requirement. ToolBench~\cite{xu2023tool,qin2023tool} needs the agent to select proper APIs from thousands of candidate APIs.

\textbf{Robustness in natural language processing} Traditional natural language robustness evaluation focuses on constructing the adversarial dataset of basic natural language tasks, such as the adversarial natural inference task~\cite{nie2019adversarial} via human attacks, adversarial BLUE tasks~\cite{wang2021adversarial} via word-level, sentence-level, and human attacks, and adversarial dialogue tasks~\cite{yu2023adversarial} via question and dialogue history Attack. Then they analyze models' (e.g., RoBERT~\cite{liu2019roberta}) behavior on these datasets. Recent robustness evaluations for LLMs try to measure their robustness to LLM's version~\cite{liu2023robustness}, search engine version~\cite{kasai2024realtime}, basic task's prompt~\cite{zhu2023promptbench,sun2023safety,hu2024prompt} and specific adversarial samples~\cite{wang2023robustness,wang2023decodingtrust,wang2023large}. 
\section{PPTC-R Benchmark}
In this section, we introduce our PowerPoint Task Completion-Robustness (PPTC-R) benchmark, including its dataset components, design principles, and the collection and validation process. 
\subsection{Introduction of the PowerPoint Task Completion benchmark}
We construct our robustness benchmark based on the open-source Powerpoint Task Completion benchmark and use its dataset, PPT tasks, and evaluation system. Here is a brief introduction to it.

\textbf{Dataset}: PPTC simulates a multi-turn dialogue between the user and the LLM, comprising 279 multi-turn sessions. Each turn within a session has a user instruction, a feasible API sequence for the instruction, and the labeled PPT file representing the correct result. To help the LLM finish the PPT task, it also provides an API reference file that contains all feasible APIs along with their description for reference. Furthermore, there's a PPT reader function that transforms the PPT file into a text-format PPT file content, as well as an API executor that executes the LLM's generated API sequence to produce the PPT prediction file.  

\textbf{PPT Task description} PPTC considers both creating new slides and editing existing PPT template tasks. Each task has its own set of sessions. To finish one turn's instruction in a session, we follow PPTC and prompt the LLM with the current instruction, previous instructions (dialogue history), the PPT file content, and the reference API file to generate an API sequence as the solution. Then the executor executes the API sequence to produce the prediction file.

\textbf{Evaluation system} We use the PPTX-evaluation system within PPTC to evaluate the correctness of the LLM prediction file. The system assesses if the objects and their position relations in the prediction file match those in the label PPT file. 

\begin{figure*}
\centering
\tcbset{colback=white!15!white, colframe=green!55!blue, fonttitle=\bfseries}
\begin{tcolorbox}[title=Sentence-level perturbation \qquad\qquad\qquad\qquad Semantic-level perturbation,sidebyside, sidebyside align=top, coltitle=black,]
\textbf{Prompt for generating irrelevant chitchat sentences:}\\Add 1$\sim$3 irrelevant chitchat non-instruction sentences into the following instruction. Don't add a new question. Instruction:<original instruction>.
\\\\
\tcblower
\textbf{Prompt for paraphrasing instructions:}
\\Rephrase the following instruction into <number> different ways: <original instruction>.
\end{tcolorbox}
\captionof{figure}{The prompts we used to create the sentence and semantic level perturbations. '<number>' is the number of paraphrased Instructions.}
\label{robust_prompt}
\end{figure*}
\subsection{Design principles}
The construction of adversarial user instructions aims to simulate possible
perturbations that naturally occur in real task-completion situations. Thus we follow three principles to construct our robustness benchmark: (1) \textbf{Realistic}: We only consider the common and daily
perturbations in the real world. For example, testing LLMs in reversed user instructions may be interesting, but this situation is impossible. So we do not consider it. (2) \textbf{Preserve semantic integrity}. We don't consider the perturbation that would change the original semantic of the instruction (e.g., deleting some sentences of the instruction randomly). We also should not add new instructions to PPTC (3) \textbf{Diverse} We should try our best to create various perturbations, making the smart LLM can not solve them by finding some simple rules. 
\subsection{Dataset collection and validation}
We construct our adversarial instructions for three levels: sentence, semantic, and language levels. We do not consider character and word level perturbations as~\cite{zhu2023promptbench} has shown the LLM's strong robustness to these simple manipulations.
\begin{itemize}
\item \textbf{Sentence-level perturbation:} We add irrelevant chitchat sentences to the original user instruction in an attempt to confuse the LLM's understanding. Specifically, for each instruction, we first prompt GPT-4 to generate 1$\sim$3 chitchat sentences, such as ‘Hey there! I hope you're having a great day. It's pretty amazing how colors can make a presentation more engaging, right?’, then we incorporate these sentences around the original user instruction (see the left part of Figure~\ref{robust_prompt}). The LLM needs to complete the user instruction while ignoring chitchat sentences. The semantics of these instructions are not changed. We further compare our perturbation approach with traditional sentence perturbation in Sec.~\ref{sec.sentence}. 

\item \textbf{Semantic-level perturbation:} For each original instruction, we prompt GPT-4 to paraphrase it in four different expressions (see the right part of Figure~\ref{robust_prompt}) and then we use the paraphrased instructions to test
the LLM’s performance. Finally, we report the average performance of the LLM in completing the instruction across four different expressions. We maintain the semantics of these instructions with various expressions. 

\item \textbf{Language-level perturbation:} To test the LLM's ability to finish the user instruction written in non-English languages, we follow the dataset XNLI~\cite{conneau2018xnli} and choose French, Spanish, German, Greek, Bulgarian, Russian, Turkish, Arabic, Vietnamese, Thai, Chinese, Hindi, Swahili and Urdu as the 14 non-English target languages. Then we use the Google Translation API to translate all user instructions from English into these target languages. We also translate the text input content in the feasible API sequence (e.g., translate Insert$\_$text ('Hello!') (English) to Insert$\_$text('{Hallo!}') (German)). Then we execute the translated feasible API sequence to obtain the label file in the target language setting. The translation operation maintains the original semantics of these instructions while expressing them in various languages. 

\end{itemize}

The change in software version usually influences the functions it can provide, and the functions can be simplified as the APIs. Thus we consider two API number perturbations to measure the LLM's robustness to the software version: 

\begin{itemize} \item \textbf{API update perturbation:} To simulate the version update scenario, we
introduce 97 new APIs along with their descriptions
into the existing API file while keeping all previous APIs unchanged\footnote{We put the new APIs in the supplementary}. These new APIs are selected from the Powerpoint keyboard and are not necessary for finishing original user instructions. But it may impact the LLM's API selection. We set the
execution result of these new APIs in the API executor as 
inserting a meaningless string '@@@'. So calling them would lead to the wrong prediction. 

\item \textbf{API lack perturbation:} To simulate the earlier software environment where some advanced functions are not provided, we only provide 24 (original 49-> now 24) basic PowerPoint APIs and the "Seek for assistance" API to the LLM. That means some parts of the user instruction may remain unfinished with the provided APIs. When the LLM finds that one part of the instruction can not be solved, it needs to call the "Seek for assistance" API to bypass this part. When it finds one part of the instruction can be solved, it needs to call the corresponding correct APIs. In nature, our objective is to measure LLMs' ability to identify whether they can complete one part of the instruction and call the correct APIs for the given situation. We set the 
execution result of the API "Seek for assistance" in the API executor as 
empty. For the label file in this perturbation, we first filter
the APIs that are in the feasible API sequence but not in the 24 basic APIs. Then we execute the filtered API sequence to obtain the
label file. We list these APIs in Appendix~\ref{appendix.lack_apis}. 
\end{itemize}

We separately conduct these five perturbations on the original PPTC benchmark to construct five different robustness settings. Our robustness benchmark consists of these five robustness settings. 
Note that our sentence and semantic perturbation approach can online generate rich adversarial data for robustness tests. Then the LLM can not improve the robustness performance cheatingly by pre-training on these adversarial data. 

\textbf{Validation} To guarantee the quality of our robustness benchmark, three of the authors check if each adversarial instruction follows the design principles. If the paraphrased instruction or the translated instruction changes the original meaning and thus violates the second principle, we discard the instruction and regenerate or use Bing to translate the original instruction until the paraphrased/translated instruction maintains semantic integrity. If the chitchat sentence contains new PPT task instruction and thus violates the second principle, we discard it and re-generate chitchat sentences. Also, if the paraphrased instruction is too similar to other paraphrased instructions, we re-paraphrase it.  


\section{Experiments}
\begin{table*}[h]
\centering
\scalebox{0.625}{
\begin{tabular}{c|ccc|ccc|ccc|ccc}
\hline
\multirow{1}*{}&\multicolumn{6}{c}{Creating new slides}&\multicolumn{6}{c}{Editing PPT template}\\
\cline{2-13}
{Models}&\multicolumn{3}{c|}{Turn-based}&\multicolumn{3}{c|}{Session-based}&\multicolumn{3}{c|}{Turn-based}&\multicolumn{3}{c}{Session-based}\\
\cline{2-13}
{}&\multicolumn{1}{c}{Original}&\multicolumn{1}{c}{Sentence}&\multicolumn{1}{c|}{Semantic}&\multicolumn{1}{c}{Original}&\multicolumn{1}{c}{Sentence}&\multicolumn{1}{c|}{Semantic}&\multicolumn{1}{c}{Original}&\multicolumn{1}{c}{Sentence}&\multicolumn{1}{c|}{Semantic}&\multicolumn{1}{c}{Original}&\multicolumn{1}{c}{Sentence}&\multicolumn{1}{c}{Semantic}\\
\hline
Davinci-003 &72.6 & 64.8~(-7.8) & 67.4~(-5.2)  &12.7 & 11.7~(-1.0) & 9.5~(-3.2) &24.4 & 26.3~(+1.9) & 25.8~(+1.4) &4.0& 0.0~(-4.0)&0.5~(-3.5)  \\
\hline
ChatGPT &70.6 & 61.3~(-9.3) & 65.0~(-5.6)  &12.7& 9.7~(-3.0) & 8.7~(-4.0) &26.3  & 28.8~(+2.5) & 27.0~(+0.3) &2.0& 2.0~(+0.0)&2.0~(+0.0)  \\
\hline
GPT-4 &75.1 & 72.3~(-2.8) & 72.0~(-3.1)  &22.7 & 12.3~(-10.4) & 14.2~(-8.5) &38.1  & 36.9~(-1.2)  & 35.8~(-6.3) &6.0& 4.0~(-2.0)&4.0~(-2.0)  \\
\hline
LLaMa-2 &16.4 & 16.3~(-0.1) & 16.1~(-0.3) &3.4& 1.7~(-1.7) & 1.0~(-2.4)  & 8.8 & 8.8~(+0.0) & 7.6~(-1.2)&0.0 & 2.0~(+2.0) &0.0~(+0.0) \\
\hline
WizardLM &23.9 & 23.8~(-0.1) & 23.8~(-0.1) &4.3& 1.0~(-3.3) & 0.0~(-4.3)  & 10.0 & 10.0~(+0.0) & 10.0~(+0.0) &0.0& 0.0~(+0.0)&0.0~(+0.0)  \\
\hline
Baichuan &15.5 & 15.5~(+0.0) & 15.0~(-0.5)  &0.0 & 1.4~(+1.4) & 1.7~(+1.7)& 4.3 & 4.3~(+0.0) & 2.5~(-1.8)  &0.0& 0.0~(+0.0)&0.0~(+0.0) \\
\hline
CodeLLaMa &36.8 & 36.2~(-0.6) & 36.8~(-0.0)  &0.0 & 0.0~(+0.0) & 0.0~(+0.0)& 18.7 & 18.8~(+0.1) & 18.7~(+0.0)  &2.0& 2.0~(+0.0)&0.0~(-2.0) \\
\hline
\end{tabular}
}
\caption{We report the robustness results of LLMs in the sentence-level and semantic-level settings in this table. 'Original' is the original accuracy copied from the PPTC benchmark. 'Sentence' and 'Semantic' are the LLM's accuracy in the sentence-level and semantic-level settings, respectively. The value in '()' is the range of change from the original performance to the robustness performance. }
\label{tab.instruction_robustness}
\end{table*}
\subsection{Large Language Models Selected for Evaluation}
We follow PPTC~\cite{Guo2023PPTCBE} and select 3 closed-source LLMs: \textit{GPT-4}~\cite{openai2023gpt4}, ~\textit{ChatGPT}, \textit{Text-Davinci-003} and 4 strong open-source LLMs: \textit{LLaMa-2-Chat}~\cite{touvron2023llama}, ~\textit{Code-LLaMa-instruct}~\cite{chiang2023vicuna}, ~\textit{WizardLM v1.2}~\cite{xu2023wizardlm}, and \textit{Baichuan-2-Chat}~\cite{yang2023baichuan} as our LLMs for evaluation. We select them as they have shown strong performance on the original PPTC benchmark or they are typical LLMs (e.g., LLaMa-2 series). For open-source LLMs, we use their chat/instruct version with 13 billion parameters. 

\begin{figure*}[ht]
  \centering
  \includegraphics[width=1\textwidth]{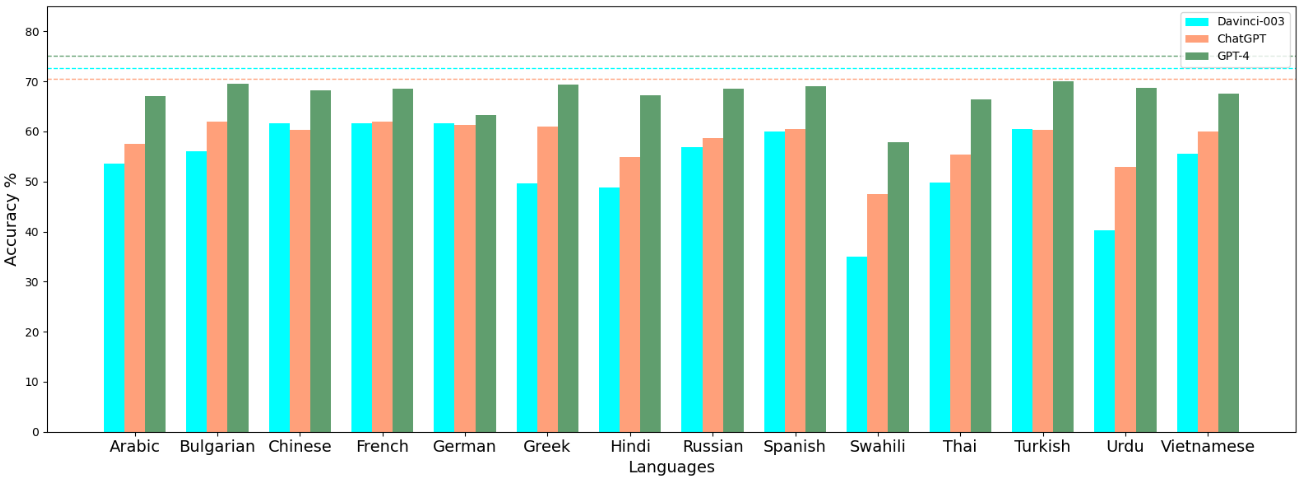}
  \caption{We illustrate the turn-based results of closed-source LLMs in the creating new slides task, where the instructions are translated into 14 non-English languages. The bar for each language represents the LLM's accuracy in the corresponding language setting. The dotted line is the LLM's accuracy when tested in the English setting. }
  \label{fig:turn_task1}
\vspace{-10pt}
\end{figure*}
\subsubsection{Evaluation approaches and metrics}
We follow PPTC~\cite{Guo2023PPTCBE} and use the two evaluation approaches: (1) Turn-based evaluation aims to measure the LLM's ability to finish a single turn where we assume that previous turns of this turn have been correctly finished. (2) Session-based evaluation tests the LLM's ability to finish the entire session containing multiple turns. Here we don't assume the LLM has correctly finished previous turns when it is asked to finish one turn of a session. 

\textbf{Metrics} In turn-based evaluation, we report the \text{turn-based accuracy} as the ratio of the number of successfully finished turns to the total number of turns. 
In session-based evaluation, we report the \text{session-based accuracy} as the ratio of the number of successfully finished sessions to the total number of sessions.

\subsection{Implementation Details}
For fair comparison and reproducibility, we follow PPTC and use the respective language models' API provided by Azure OpenAI Service for closed-source LLMs. For open-source LLMs, we download them from the official websites. More details are in Appendix~\ref{appendix.detail}. 
\subsection{Main results}
\label{sec.results}
In this section, we report the accuracy results of LLMs on our benchmark in Tables~\ref{tab.instruction_robustness}~\&~\ref{tab.api_robust} and Figures~\ref{fig:turn_task1}~\&~\ref{fig:sess_task1}~\&~\ref{fig:turn_task2}. The results of the cost measurement are in Appendix~\ref{appendix.results}. Then we analyze the results from the aspects of LLMs and perturbation types.

\begin{table*}[ht]
\centering
\scalebox{0.62}{
\begin{tabular}{c|ccc|ccc|ccc|ccc}
\hline
\multirow{1}*{}&\multicolumn{6}{c}{Creating new slides}&\multicolumn{6}{c}{Editing PPT template}\\
\cline{2-13}
{Models}&\multicolumn{3}{c|}{Turn-based}&\multicolumn{3}{c|}{Session-based}&\multicolumn{3}{c|}{Turn-based}&\multicolumn{3}{c}{Session-based}\\
\cline{2-13}
{}&\multicolumn{1}{c}{Original}&\multicolumn{1}{c}{Lack}&\multicolumn{1}{c|}{Update}&\multicolumn{1}{c}{Original}&\multicolumn{1}{c}{Lack}&\multicolumn{1}{c|}{Update}&\multicolumn{1}{c}{Original}&\multicolumn{1}{c}{Lack}&\multicolumn{1}{c|}{Update}&\multicolumn{1}{c}{Original}&\multicolumn{1}{c}{Lack}&\multicolumn{1}{c}{Update}\\
\hline
Davinci-003 &72.6 & 55.1~(-17.5) & 44.5~(-28.1)  &12.7 & 5.2~(-7.5) & 1.3~(-11.4)&24.4 & 33.7~(+9.3) & 17.5~(-6.9) &4.0& 0.0~(-4.0)&0.0~(-4.0)  \\
\hline
ChatGPT &70.6 & 55.4 (-15.2) & 55.4~(-15.2)  &12.7& 3.9~(-8.8) & 5.3~(-7.4) &26.3  & 27.5~(+1.2) & 15.0~(-11.3) &2.0& 0.0~(-2.0)&0.0~(-2.0)  \\
\hline
GPT-4 &75.1 & 62.5 (-12.6)& 75.7~(+0.6)  &22.7 & 5.2~(-17.5) & 18.8~(-3.9) &38.1  & 39.4~(+1.3)  & 35.6~(-2.5) &6.0& 0.0~(-6.0)&2.0~(-4.0)  \\
\hline
LLaMa-2 &16.4 & 16.5 (+0.1)& 7.8 (-8.6)& 3.4&0.0~(-3.4) & 3.4~(-0.0)  & 8.8 & 12.0~(+4.0) & 7.5~(-1.3)&0.0 & 0.0~(+0.0) &2.0~(+2.0) \\
\hline
WizardLM &23.9 & 18.9 (-5.0)& 11.3 (-12.6)  & 4.3&0.0~(-4.3) & 0.0~(-4.3)  & 10.0 & 14.4~(+4.4) & 6.9~(-3.1) &0.0& 0.0~(+0.0)&0.0~(+0.0)  \\
\hline
Baichuan &15.5 & 18.7 (+3.2)& 13.2 (-2.3)  &0.0 & 0.0~(+0.0) & 1.0~(+1.0)  & 4.3 & 10.6~(+6.3) & 2.5~(+1.8) &0.0& 6.0~(+6.0)&0.0~(+0.0) \\
\hline
CodeLLaMa &36.8 & 26.3~(-10.5) & 22.4~(-14.4)  &0.0 & 0.0~(+0.0) & 1.0~(+2.0)& 18.7 & 13.6~(-5.1) & 12.6~(-6.1)  &2.0& 2.0~(+0.0)&2.0~(+0.0) \\
\hline
\end{tabular}
}
\caption{We report the robustness results of LLMs in the API-lack and API-update settings in this table. 'Original' is the original accuracy copied from the PPTC benchmark. 'Lack' and 'Update' are the LLM's accuracy in the API-lack and API-update settings, respectively.}
\label{tab.api_robust}
\end{table*}
\begin{table}[ht]
\centering
\scalebox{0.625}{
\begin{tabular}{c|cc|c}
\hline
{Models}&\multicolumn{2}{c|}{Creating new slides}&\multicolumn{1}{c}{Editing PPT template}\\
\cline{2-4}
{}&\multicolumn{1}{c}{Turn-based}&\multicolumn{1}{c}{Session-based}&\multicolumn{1}{|c}{Turn-based}\\
\hline
ChatGPT &16.3 & 54.7 & 19.7 \\
\hline
GPT-4 &9.4 & 54.1 & 17.7 \\
\hline
LLaMa-2 &13.6 & 55.2 & 9.4\\
\hline
WizardLM &18.5 & 94.2 & 10.3 \\
\hline
\end{tabular}
}
\caption{This table presents Average Performance Drop Rate (APDR) results for LLMs. For the "creating new slides" task's turn-based column, we compute LLMs' PDR rates using their turn-based accuracy of the creating new slides task in each robustness setting. The average is reported as APDR. The same calculation is applied to the other columns. Note that we exclude the multilingual setting for open-source LLMs.}
\label{tab.APDR}
\end{table}

\begin{table*}[h]
\centering
\scalebox{0.625}{
\begin{tabular}{c|ccc|ccc|ccc|ccc}
\hline
\multirow{1}*{}&\multicolumn{6}{c}{Creating new slides}&\multicolumn{6}{c}{Editing PPT template}\\
\cline{2-13}
{Models}&\multicolumn{3}{c|}{Turn-based}&\multicolumn{3}{c|}{Session-based}&\multicolumn{3}{c|}{Turn-based}&\multicolumn{3}{c}{Session-based}\\
\cline{2-13}
{}&\multicolumn{1}{c}{Original}&\multicolumn{1}{c}{ChitChat}&\multicolumn{1}{c|}{True or False}&\multicolumn{1}{c}{Original}&\multicolumn{1}{c}{ChitChat}&\multicolumn{1}{c|}{True or False}&\multicolumn{1}{c}{Original}&\multicolumn{1}{c}{ChitChat}&\multicolumn{1}{c|}{True or False}&\multicolumn{1}{c}{Original}&\multicolumn{1}{c}{ChitChat}&\multicolumn{1}{c}{True or False}\\
\hline
GPT-4 &75.1 & 72.3~(-2.8) & 73.2~(-1.9)  &22.7 & 12.3~(-10.4) & 14.8~(-7.9) &38.1  & 36.9~(-1.2)  & 37.5~(-0.6) &6.0& 4.0~(-2.0)&6.0~(+0.0)  \\
\hline
\end{tabular}
}
\caption{We report the robustness results of GPT-4 in the sentence-level setting by adding chitchat sentences ('Chitchat') and randomly generated strings ('Random'), respectively.}
\label{tab.vs}
\end{table*}
\textbf{Sentence-level and semantic-level robustness:} We report LLMs' robustness performance for Sentence-level and semantic-level settings in Table~\ref{tab.instruction_robustness}.
From the results, we highlight the following key findings: (1) For both two tasks, GPT-4 shows the strongest performance under the sentence-level and semantic-level perturbations and it usually drops its performance less than other closed-source LLMs. (2) Open-source LLMs achieve low performance but also drop less performance than their closed-source counterparts. CodeLLaMa achieves the strongest robustness and performance among open-source LLMs. (3) Overcoming the sentence-level perturbation is harder than the semantic-level perturbation as LLMs usually drop much performance in the former setting for both the turn-based and session-based evaluation. 

\textbf{Language-level robustness:} We illustrate the turn-based results of closed-source LLMs\footnote{Current open-source LLMs claim that they are mainly pre-trained in English corpus (e.g., only 1\% non-English corpus for Llama-2) and do not support multi-lingual settings.} in the language-level robustness setting in Figure~\ref{fig:turn_task1}. Due to space limitation, we put other results of the multilingual setting in Appendix~\ref{appendix.multilingual}. From these results, we find that (1) GPT-4 outperforms other LLMs obviously in the turn-based evaluation (see Figure~\ref{fig:turn_task1} and ~\ref{fig:turn_task2}) and also drops less than the other two LLMs, which shows GPT-4's strong multilingual understanding ability. (2) Even GPT-4 also performs poorly in low-resource languages like Swahili. ChatGPT and Davinci further perform poorly in Urdu and Arabic. That means improving LLM's performance in low-resource languages is still a long-term challenge. 

\textbf{API lack and API update robustness:} We report the results of the API lack and update settings in Table~\ref{tab.api_robust}. From the results, we find that:(1) GPT-4 shows good robustness to the API-update perturbation as it only drops the performance slightly by 2$\sim$4 percent. In contrast, other closed-LLMs drop their performance markedly as the perturbation of introducing more new APIs. That shows the unique position of GPT-4 in API calls. (2) For the creating slides task, LLMs drop more performance in the API-lack setting than they in the API-update setting. That means it is harder for current LLMs to know what they can not do based on the provided API list. We analyze the wrong examples and find that is because LLMs call unavailable APIs rather than seek assistance. (3) Surprisingly, we observe that the turn-based performance of almost all LLMs in the editing task improves under the API-lack perturbation. We discover that this is attributed to the editing task design in PPTC, which emphasizes instructing LLMs to process lengthy PPT templates with high-frequency basic APIs. Consequently, the API-lack setting occasionally reduces the task difficulty by deleting low-frequency APIs. 

\textbf{LLMs' robustness varies with the difficulty of the task and the complexity of the environment}
To measure the variation of LLMs' robustness, we use the Average Performance Drop Rate (APDR) metric proposed in~\cite{zhu2023promptbench}\footnote{We don't use this rate to compare different LLMs as open-source LLMs can achieve low APDR rates with a pretty low performance. Then the comparison is useless.}, which quantifies the average relative performance decline of LLMs when subjected to perturbation attacks\footnote{It is calculated by dividing the range of performance variation with the original performance.}. We report the APDR rate of LLMs in Table~\ref{tab.APDR}. We find a decrease in the robustness of LLMs when we increase the difficulty of the same task (e.g., turn-based evaluation->session-based evaluation) or move to complex environments (e.g., creating 1$\sim$2 slides->editing the long template with tens of slides), even for GPT-4. For instance, in the task of creating new slides, GPT-4's APDR rate increases from 9.4 to 54.1 (where a higher value indicates poorer robustness) when transitioning from turn-based to session-based task evaluation. The table further indicates that completing multi-turns (session-based evaluation) is harder than editing the template as LLMs exhibit higher APDR rates in the former, and occasionally, open-source LLMs' APDR rate experiences a slight drop in the latter. 
\begin{figure*}[ht]
  \centering
  \includegraphics[width=1\textwidth]{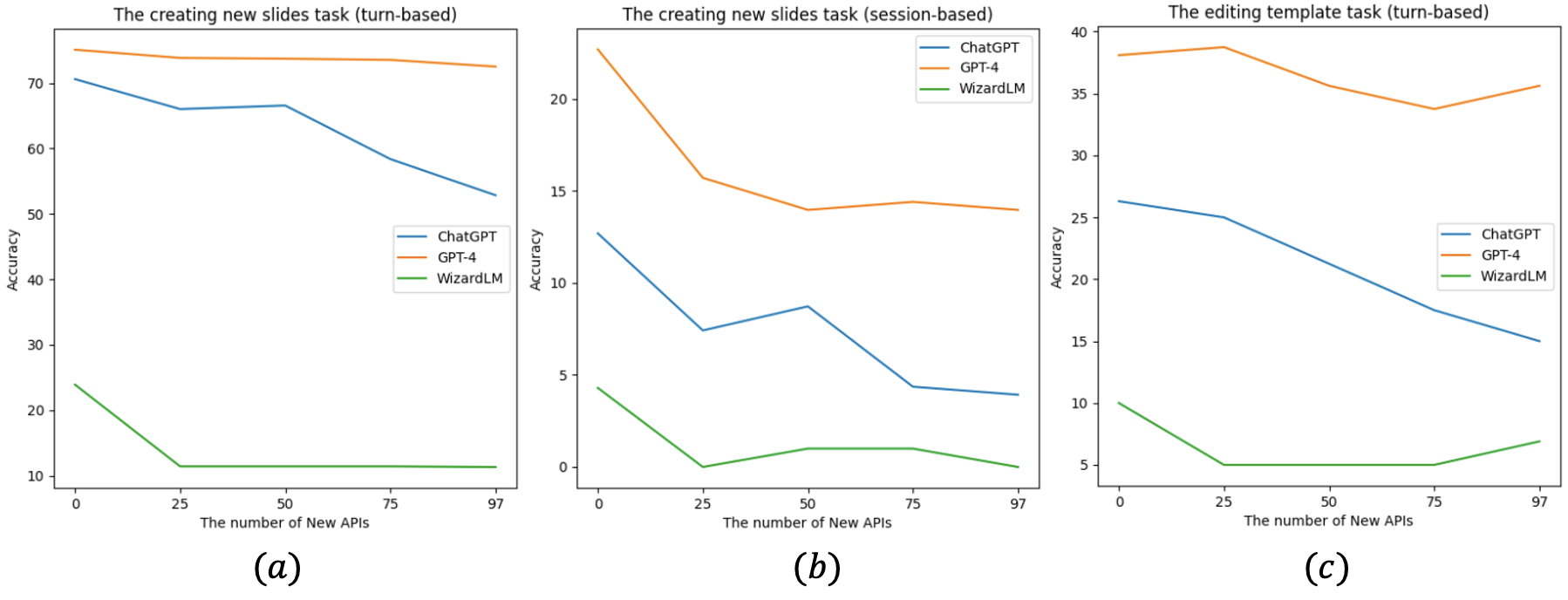}
  \caption{We report the results of three LLMs with different numbers of new APIs in sub-figures (a), (b), and (c). The session-based accuracy of all LLMs' editing template task performance is pretty low (<4).}
  \label{fig:api_Number}
\vspace{-12pt}
\end{figure*}
\section{Analysis}
\label{sec.ana}
\subsection{Error analysis for LLMs}
\label{sec.error}
To conduct error analysis in PPTC-R, for each robustness setting, we randomly collect 25 examples that are wrong in PPTC-R but are correct in the original PPTC. We do the collection process separately for ChatGPT and GPT-4. Based on these collected examples, we identify the following error reasons: (1) Being Distracted by chitchat sentences. When processing the user instruction in the sentence-level robustness setting, LLMs start to chat with the user and forget to generate API sequences, with a rate of 61\% in all errors. (2) Calling unavailable APIs or new APIs. In the API lack and multilingual settings, we find that GPT-4 and ChatGPT tend to create new APIs like 'select$\_$column'. Though these APIs may appear reliable, they are non-executable. In the API update setting, ChatGPT invokes new APIs provided by our setting. Although these APIs are executable, they are not necessary. (3) Misunderstanding the instruction. In the sentence and semantic-level robustness settings, GPT-4 and ChatGPT might misunderstand instructions, leading to the call of wrong APIs. We provide detailed examples in Appendix~\ref{appendix,wrong_example}. 
\subsection{Chitchat sentences Vs 'True or False'}
~\label{sec.sentence}
Traditional sentence-level perturbations are typically performed by inserting 'True or False' (StressTest~\cite{naik2018stress}) or a randomly generated string like 'KjPJJ2a7RB' (Checklist~\cite{ribeiro2020beyond}) into the original input. In contrast, our sentence-level perturbation is more challenging as it introduces various chitchat sentences. Empirically, we test GPT-4 in a robustness setting where we prepend 'True or False' to each instruction. From Table~\ref{tab.vs}, we observe that our perturbation indeed causes GPT-4 to experience a greater performance drop, as it deviates from chitchat. 
\subsection{\textit{How does the number of new APIs influence the LLM's performance?}}
\label{sec.api_number}
We study the influence of different numbers of new APIs on LLMs' performance in Figure~\ref{fig:api_Number}. We find that ChatGPT's performance gradually decreases when increasing the number of APIs. The performance of open-source WizardLM first quickly drops when we add 25 new APIs to its API reference list and then maintains its performance at a pretty low level. This means current most LLMs still lack enough robustness to select the correct APIs from a larger candidate pool (version update). In contrast, GPT-4 maintains its turn-based performance at a comparatively high level, which shows its strong robustness to the API update setting. However, its session-based performance in the creating task also drops obviously as the double challenges of increasing the new API number and finishing the entire multi-turn session.
\section{Conclusion}
Deploying LLMs and LLM-based agents to complete users' task instructions has become a pressing demand. However, there still lacks a evaluation and analysis of LLMs' robustness in complex task completions. We introduce the PowerPoint Task Completion-Robustness benchmark, designed to assess LLMs' robustness in handling user adversarial instructions and adapting to different software versions in complex PPT task completion. The results of 7 LLMs in our benchmark show that GPT-4 is the strongest one but all LLMs' robustness degrades when increasing the task difficulty. We further conduct a detailed analysis of error reasons and robustness behaviors for profound understanding.
\section{Limitations and potential risks}
Investigating the LLM's robustness to the PPT file (environment) may be interesting. A simple way is to vary the number of shapes in the PPT file. For example, slides containing more figures may pose a greater challenge for LLMs when completing figure-related instructions. However, we do not consider this perturbation as it is hard to control in specific slides. For example, some slides may allow the addition of more figures, while others can not as they are completely fulfilled. On the other hand, our benchmark does not consider creating harder instructions by further asking experts to write and edit the instructions. But current LLMs have already dropped their performance obviously in our setting. So we leave the further creation work in the future.   

We do not see any potential physical risk in our benchmark as we just test the LLM's robustness to do virtual PPT tasks under perturbations. We also do not see any societal risk.
\bibliography{anthology,custom}

\appendix
\section{Basic APIs in the API lack  setting}
\label{appendix.lack_apis}
We list the APIs used in the API lack setting in Figure~\ref{fig.api_lack_1}~$\&$~\ref{fig.api_lack_2}. We select them as they provide basic functions in PowerPoint software with high usage frequency in the benchmark. 
\begin{figure*}
\tcbset{colback=green!-15!white, colframe=green!65! black, fonttitle=\bfseries}
\begin{tcolorbox}[title=API reference file, sidebyside align=top]
API: create$\_$slide(): This API creates a new slide.

API: set$\_$background$\_$color(color): This API sets the background color of the slide.
It takes one parameter 'color', the color name to set as a string, such as 'red', 'purple'.

API: choose$\_$title(): This API selects the title on the slide.
You should first call choose$\_$title() before inserting text to or changing font attributes of the title.

API: choose$\_$content(): This API select the content on the slide.
You should first call choose$\_$content() before inserting text to or changing font attributes of the content.

API: choose$\_$textbox(idx): This API selects the textbox element on the slide.
It takes one parameter, the index of textbox as integer. idx is set to 0 by default, meaning the first textbox.
You should first call choose$\_$textbox() before inserting text to or changing font attributes of the textbox element.

API: choose$\_$picture(idx): This API selects the picture element on the slide.
It takes one parameter, the index of textbox as integer. idx is set to 0 by default, meaning the first textbox.
You should first call choose$\_$picture() before changing height, width, rotation of the picture element. You should not call choose$\_$picture() before inserting picture element.

API: choose$\_$shape(shape$\_$name): This API selects a specific shape by shape name on the slide.
It takes one parameter 'shape$\_$name', the name of the shape to select as a string.         shape$\_$name can be chosen from ['rectangle','right$\_$arrow','rounded$\_$rectangle','triangle','callout','cloud','star','circle']
You should first call choose$\_$shape(shape$\_$name) before you can do operations on the shape. You should not call choose$\_$shape(shape$\_$name) before inserting shape element.

API: choose$\_$table(): This API selects the table element on the slide.
You should first call choose$\_$table() before changing the table. You should not call choose$\_$table() before inserting table element.

API: choose$\_$table$\_$cell(row$\_$id, column$\_$id): This API selects a specific cell in the table by giving row$\_$id and column$\_$id.
It takes two parameters, the row id and column id of the cell to select as integers (id starts from 0). Remember the first parameter is row id, the second parameter is column id.
You should first call choose$\_$table$\_$cell(row$\_$id, column$\_$id) before inserting text into a specific cell of the table.

API: set$\_$width(width): This API sets the width of the selected object.
It takes one parameter 'width', the width of an object in centimeters as float.
You should first choose an object before you can change the width of it.

API: set$\_$height(height): This API sets the height of the selected object.
It takes one parameter 'height', the height of an object in centimeters as float.
You should first choose an object before you can change the height of it

API: set$\_$left(left): This API moves and changes the object's position. It sets the x position of the selected object's leftmost point.
It takes one parameter, the x position to set.
You should first choose an object before you can change the left of it

API: set$\_$top(top): This API moves and changes the object's position. It sets the y position of the selected object's upmost point.
It takes one parameter, the y position to set.
You should first choose an object before you can change the top of it

API: insert$\_$text(text): This API inserts text into a text frame (textbox, title, content, table).

API: set$\_$font$\_$size(font$\_$size): This API sets the size of the font
It can take one argument 'font$\_$size', the font size to set as an integer.

API: set$\_$font$\_$color(color): This API sets the color of the font.
It takes one parameter 'color', the color name to set as a string, such as 'red', 'purple'.

API: set$\_$font$\_$bold(): This API sets the font to be bold.
\end{tcolorbox}
\captionof{figure}{The reference API file in the API-lack setting.}
\label{fig.api_lack_1}
\vspace{-5pt}
\end{figure*}

\begin{figure*}
\tcbset{colback=green!-15!white, colframe=green!65! black, fonttitle=\bfseries}
\begin{tcolorbox}[title=API reference file, sidebyside align=top]
API: insert$\_$picture(picture$\_$name): This API inserts a picture onto the slide.
It takes one parameter 'picture$\_$name', the name or description of picture as a string

API: insert$\_$rectangle(): This API inserts a rectangle or square shape onto the slide.

API: insert$\_$right$\_$arrow(): This API inserts an arrow shape onto the slide.

API: insert$\_$table(row$\_$num, col$\_$num): This API inserts a table of row$\_$num rows and col$\_$num columns onto the current slide.
It takes two argument, the row number and the column number of the inserted table as integer. Remember the first parameter is row number and the second parameter is column number.

API: insert$\_$line$\_$chart(data, series): This API inserts a line chart onto the slide.
It takes two argument, 'data' is a list of numbers and 'series' is a list of strings.

API: insert$\_$bar$\_$chart(data, series): This API inserts a bar chart onto the slide.
It takes two argument, 'data' is a list of numbers and 'series' is a list of strings.

API: insert$\_$pie$\_$chart(data, series): This API inserts a pie chart onto the slide.
It takes two argument, 'data' is a list of numbers and 'series' is a list of strings.

API: seek$\_$assistance(): This API requests human help when the computer is unsure about the result or lacks the necessary API to fulfill the user's instruction.
\end{tcolorbox}
\captionof{figure}{The reference API file in the API-lack setting.}
\label{fig.api_lack_2}
\vspace{-5pt}
\end{figure*}

\section{Experimental details}
\label{appendix.detail}
For closed-source LLMs, Azure OpenAI services\footnote{\url{https://azure.microsoft.com/en-us/products/cognitive-services/openai-service}} offer two API types: completion and chat completion. Completion API generates text from prompts, while chat completion API responds based on conversation history and new input. We use the completion API for Text-Davinci-003 and the chat completion API for ChatGPT and GPT-4. We set a temperature of zero for deterministic output and a max token limit of 2048. Frequency penalty and top p are kept at their default values of zero and 1, respectively. For open-source LLMs, we choose the chat version of Llama-2, the v1.2 version of WizardLM, and the chat version of Baichuan as our open-source LLMs. We choose the 13 billion parameters model of the three LLMs. If the token number of the input prompt is beyond the token limit, we cut the PPT content to reduce the token number of the prompt.

The inference prompts in the turn-based evaluation and session-based evaluation have two differences: the API solutions for previous turns in dialogue history are the correct API sequences in the turn-based evaluation and the outputs of the LLM in the session-based evaluation. (2) The PPT content is parsed from the PPT file. The PPT file is obtained by executing the label API sequences in the turn-based evaluation and the previous outputs of the LLM in the session-based evaluation. That means the error made by LLMs in previous turns would influence subsequent turns in the session-based evaluation. 
We copy the inference prompt we used from PPTC and illustrate it in Figure~\ref{inference_prompt}.
\begin{figure*}
\tcbset{colback=green!-15!white, colframe=green!65! black, fonttitle=\bfseries}
\begin{tcolorbox}[title=Inference prompt in PPTC, sidebyside align=top]
\small{(\textbf{Task instruction}) You are an AI assistant to help the user to operate PowerPoint and edit the contents.\\}
\small{ Give you the user instruction:<Current user instruction>, you can complete it based on the following APIs and PPT file content. Current you are at page <Page id>. Please finish the user instruction with the functions you have.
Don't generate instructions beyond what the user has instructed. 
Don't guess what the user may instruct in the next step and generete API for them.
Don't use python loop to call API. You can only call API once in one line.
If the user does not specify the page to be modified, you can directly start using the APIs without having to navigate to other pages.

You need to generate code which can finish the user instruction. The multiple lines of code should be surrounded by <code> and </code> such as:
<code>
API();
API();
</code>

For example, if the user instruction is "create a slide", then the answer should be:
\\<code>
create\_slide();
</code>}
\\\\\small{(\textbf{API file}) Now, you have access to a list of PowerPoint APIs with the following functions: <APIs and their descriptions> \\(e.g.,API(name="set$\_$width", parameters="(width)", \\description="This API sets the width of the selected object.",
        \\parameter$\_$description="It takes one parameter 'width', the width of an object in centimeters as float.",
        \\composition$\_$instruction="You should first choose an object before you can change the width of it.",\\api$\_$desc="width of picture and shapes")
)
\\\\(\textbf{PPT file content}) All the PPT contents are:
\\<Begin of PPT>
\\\textit{Turn-based: <Parsed PPT file content of the label PPT file of the previous turns>\\Session-based: <Parsed PPT file content of the LLM prediction file of the previous turns>}\\<End of PPT>
\\\\\small{(\textbf{Dialogue history}) 
\\¬User¬:
Hello!
\\¬AI¬:
Hi there! How can I help you?
\\¬User¬:
<the first instruction>
\\¬AI¬:
\\\textit{Turn-based: <the correct feasible API sequence>,
\\Session-based: <the LLM-generated API sequence>}
\\...
\\¬User¬:
<Current user instruction>. Surrounding your answer with <code> and </code>.
\\¬AI¬:}}
\end{tcolorbox}
\captionof{figure}{The inference prompt we used in both turn-based and session-based evaluation settings. }
\label{inference_prompt}
\end{figure*}

\section{Detailed Results of LLMs on PPTC-R benchmark}
In turn-based evaluation, we report the average token number of the input of one turn and the average API number for finishing one turn as the cost measurement.
In session-based evaluation, we report the average value of the token number of all inputs in one session and the average API number required to complete one session as the cost measurement.
We return the accuracy and the cost measurement in both two evaluations in Table~\ref{tab.noisy},~\ref{tab.robust},~\ref{tab.API_lack}, and ~\ref{tab.API_update}. 
\begin{table*}[h]
\centering
\scalebox{0.6}{
\begin{tabular}{c|ccc|ccc|ccc|ccc}
\hline
\multirow{3}*{Models and Methods}&\multicolumn{6}{c}{Creating new slides}&\multicolumn{6}{c}{Editing PPT template}\\
\cline{2-13}
{}&\multicolumn{3}{c|}{Turn-based}&\multicolumn{3}{c|}{Session-based}&\multicolumn{3}{c|}{Turn-based}&\multicolumn{3}{c}{Session-based}\\
\cline{2-13}
{}&\multicolumn{1}{c}{Accuracy}&\multicolumn{1}{c}{Avg token}&\multicolumn{1}{c|}{Avg API}&\multicolumn{1}{c}{Accuracy}&\multicolumn{1}{c}{Avg token}&\multicolumn{1}{c|}{Avg API}&\multicolumn{1}{c}{Accuracy}&\multicolumn{1}{c}{Avg token}&\multicolumn{1}{c|}{Avg API}&\multicolumn{1}{c}{Accuracy}&\multicolumn{1}{c}{Avg token}&\multicolumn{1}{c}{Avg API}\\
\hline
Davinci-003 & 64.8 & 2872.2 & 3.1 & 11.7 & 20716.8 & 24.2 & 26.3 & 2915.8 & 8.3 & 0.0 & 9321.1 & 23.6 \\
\hline
ChatGPT & 61.3 & 3106.6 & 3.4 & 9.7 & 22611.1 & 26.0 & 28.8 & 4140.9 & 8.1 & 2.0 & 13240.0 & 26.8 \\
\hline
GPT-4 & 72.3 & 3111.2 & 3.0 & 12.3 & 22438.0 & 21.6 & 36.9 & 7565.9 & 7.7 & 4.0 & 24185.0 & 24.0 \\
\hline
LLaMa-2 & 16.3 & 2822.6 & 4.3 & 1.7 & 11018.5 & 60.3 & 8.8 & 4124.5 & 7.6 & 2.0 & 4173.0 & 15.4 \\
\hline
WizardLM & 23.8 & 1327.1 & 3.3 & 1.0 & 11494.4 & 22.8 & 10.0 & 1328.4 & 5.7 & 0.0 & 4303.7 & 9.5 \\
\hline
Baichuan & 15.5 & 1327.1 & 9.8 & 1.4 & 10548.9 & 56.1 & 4.3 & 1328.0 & 9.6 & 0.0 & 4256.4 & 25.0 \\
\hline
CodeLLaMa & 36.2 & 2814.3 & 3.5 & 0.0 & 20720.9 & 32.1 & 18.8 & 2061.7 & 7.5 & 2.0 & 9566.9 & 22.58 \\
\hline
\end{tabular}
}
\caption{We report the results of LLMs in the sentence-level robustness setting in this table.' Davinci-003' is the Text-Davinci-003 model.}
\label{tab.noisy}
\end{table*}

\begin{table*}[h!]
\centering
\scalebox{0.6}{
\begin{tabular}{c|ccc|ccc|ccc|ccc}
\hline
\multirow{3}*{Models and Methods}&\multicolumn{6}{c}{Creating new slides}&\multicolumn{6}{c}{Editing PPT template}\\
\cline{2-13}
{}&\multicolumn{3}{c|}{Turn-based}&\multicolumn{3}{c|}{Session-based}&\multicolumn{3}{c|}{Turn-based}&\multicolumn{3}{c}{Session-based}\\
\cline{2-13}
{}&\multicolumn{1}{c}{Accuracy}&\multicolumn{1}{c}{Avg token}&\multicolumn{1}{c|}{Avg API}&\multicolumn{1}{c}{Accuracy}&\multicolumn{1}{c}{Avg token}&\multicolumn{1}{c|}{Avg API}&\multicolumn{1}{c}{Accuracy}&\multicolumn{1}{c}{Avg token}&\multicolumn{1}{c|}{Avg API}&\multicolumn{1}{c}{Accuracy}&\multicolumn{1}{c}{Avg token}&\multicolumn{1}{c}{Avg API}\\
\hline
Davinci-003 & 67.4 & 2781.3 & 2.4 & 9.5 & 20065.9 & 25.0 & 25.8 & 2892.9 & 7.8 & 0.5 & 9247.9 & 23.3 \\
\hline
ChatGPT & 65.0 & 2887.1 & 3.3 & 8.7 & 20865.0 & 25.3 & 27.0 & 4127.8 & 8.1 & 2.0 & 13207.0 & 26.3 \\
\hline
GPT-4 & 72.0 & 2887.7 & 3.0 & 14.2 & 20817.7 & 22.2 & 35.8 & 7538.3 & 7.8 & 4.0 & 24103.1 & 24.4 \\
\hline
LLaMa-2 & 16.1 & 2822.6 & 4.3 & 1.0 & 9777.9 & 16.6 & 7.6 & 2983.8 & 6.4 & 0.0 & 9550.7 & 22.8 \\
\hline
WizardLM & 23.8 & 1327.1 & 3.4 & 0.0 & 11494.4 & 22.8 & 10.0 & 1328.5 & 5.8 & 0.0 & 4303.7 & 9.5 \\
\hline
Baichuan & 15.0 & 1327.1 & 10.0 & 0.0 & 10112.3 & 24.1 & 2.5 & 1328.5 & 12.2 & 0.0 & 4256.4 & 17.0 \\
\hline
CodeLLaMa & 36.8 & 2819.7 & 3.4 & 0.0 & 20720.9 & 32.1 & 18.8 & 2983.1 & 7.3 & 0.0 & 10351.1 & 25.0 \\
\hline
\end{tabular}
}
\caption{We report the results of LLMs in the semantic-level robustness setting in this table. Each result is the average performance in finishing four different paraphrased instructions.}
\label{tab.robust}
\end{table*}

\begin{table*}[h!]
\centering
\scalebox{0.6}{
\begin{tabular}{c|ccc|ccc|ccc|ccc}
\hline
\multirow{3}*{Models and Methods}&\multicolumn{6}{c}{Creating new slides}&\multicolumn{6}{c}{Editing PPT template}\\
\cline{2-13}
{}&\multicolumn{3}{c|}{Turn-based}&\multicolumn{3}{c|}{Session-based}&\multicolumn{3}{c|}{Turn-based}&\multicolumn{3}{c}{Session-based}\\
\cline{2-13}
{}&\multicolumn{1}{c}{Accuracy}&\multicolumn{1}{c}{Avg token}&\multicolumn{1}{c|}{Avg API}&\multicolumn{1}{c}{Accuracy}&\multicolumn{1}{c}{Avg token}&\multicolumn{1}{c|}{Avg API}&\multicolumn{1}{c}{Accuracy}&\multicolumn{1}{c}{Avg token}&\multicolumn{1}{c|}{Avg API}&\multicolumn{1}{c}{Accuracy}&\multicolumn{1}{c}{Avg token}&\multicolumn{1}{c}{Avg API}\\
\hline
Davinci-003 & 55.1 & 2125.0 & 3.4 & 5.2 & 15527.1 & 25.2 & 33.75 & 2720.7 & 8.0 & 0.0 & 8720.3 & 25.0 \\
\hline
ChatGPT & 55.4 & 2138.9 & 3.6 & 3.9 & 15631.6 & 26.3 & 27.5 & 3925.5 & 8.8 & 0.0 & 12567.9 & 26.9 \\
\hline
GPT-4 & 62.5 & 2138.9 & 3.0 & 5.2 & 15572.4 & 22.2 & 39.4 & 7265.1 & 7.6 & 0.0 & 23251.6 & 24.9 \\
\hline
LLaMa-2 & 16.5 & 2070.4 & 5.7 & 0.0 & 17322.6 & 49.5 & 12.0 & 2787.8 & 7.8 & 0.0 & 8993.0 & 20.8 \\
\hline
WizardLM & 18.9 & 1308.8 & 3.2 & 0.0 & 15885.4 & 121.5 & 14.4 & 1306.7 & 5.6 & 0.0 & 13508.7 & 29.5 \\
\hline
Baichuan & 18.7 & 1310.0 & 10.1 & 0.0 & 11335.2 & 66.4 & 10.6 & 1308 & 8.5 & 6.0 & 4209.6 & 22.5 \\
\hline
CodeLLaMa & 26.3 & 2061.7 & 4.4 & 0.0 & 14448.1 & 34.7 & 13.6 & 2791.8 & 7.6 & 2.0 & 10001.8 & 13.3 \\
\hline
\end{tabular}
}
\caption{We report the results of LLMs in the API lack setting in this table. In this setting, we only maintain the 24 basic APIs. LLMs only need to finish the content that can be finished by the 24 APIs.}
\label{tab.API_lack}
\end{table*}

\begin{table*}[h!]
\centering
\scalebox{0.6}{
\begin{tabular}{c|ccc|ccc|ccc|ccc}
\hline
\multirow{3}*{Models and Methods}&\multicolumn{6}{c}{Creating new slides}&\multicolumn{6}{c}{Editing PPT template}\\
\cline{2-13}
{}&\multicolumn{3}{c|}{Turn-based}&\multicolumn{3}{c|}{Session-based}&\multicolumn{3}{c|}{Turn-based}&\multicolumn{3}{c}{Session-based}\\
\cline{2-13}
{}&\multicolumn{1}{c}{Accuracy}&\multicolumn{1}{c}{Avg token}&\multicolumn{1}{c|}{Avg API}&\multicolumn{1}{c}{Accuracy}&\multicolumn{1}{c}{Avg token}&\multicolumn{1}{c|}{Avg API}&\multicolumn{1}{c}{Accuracy}&\multicolumn{1}{c}{Avg token}&\multicolumn{1}{c|}{Avg API}&\multicolumn{1}{c}{Accuracy}&\multicolumn{1}{c}{Avg token}&\multicolumn{1}{c}{Avg API}\\
\hline
Davinci-003 & 44.5 & 2938.8 & 2.8 & 1.3 & 21178.7 & 21.1 & 17.5 & 2942.6 & 6.6 & 0.0 & 9419.1 & 18.9 \\
\hline
ChatGPT & 55.4 & 4603.6 & 2.9 & 5.2 & 33166.4 & 23.2 & 15.0 & 4605.9 & 7.1 & 0.0 & 14724.6 & 20.3 \\
\hline
GPT-4 & 75.7 & 6495.9 & 2.8 & 18.8 & 46747.9 & 20.7 & 35.6 & 8511.7 & 7.5 & 2.0 & 27211.5 & 23.4 \\
\hline
LLaMa-2 & 7.8 & 2318.8 & 10.0 & 3.4 & 10073.8 & 17.2 & 7.5 & 2137.8 & 8.7 & 2.0 & 9910 & 13.7 \\
\hline
WizardLM & 11.3 & 1317.3 & 2.5 & 0.0 & 10285.4 & 11.8 & 6.9 & 1321.0 & 5.4 & 0.0 & 10406.5 & 33.3 \\
\hline
Baichuan & 13.2 & 1325.7 & 5.8 & 1.0 & 12018.5 & 60.3 & 2.5 & 1320.7 & 10.2 & 0.0 & 9818.0 & 22.8 \\
\hline
CodeLLaMA & 22.4 & 3134.6 & 2.3 & 1.0 & 22536.2 & 17.2 & 12.6 & 3137.1 & 5.4 & 2.0 & 10001.1 & 13.3 \\
\hline
\end{tabular}
}
\caption{We report the results of LLMs in the API update setting in this table. In this setting, we add 97 new APIs into the prompt to simulate the version update. }
\label{tab.API_update}
\end{table*}
\label{appendix.results}
\section{Closed-source LLM's Multilingual Results in the Editing Template Task}
\label{appendix.multilingual}
\begin{figure*}[ht]
  \centering
  \includegraphics[width=1\textwidth]{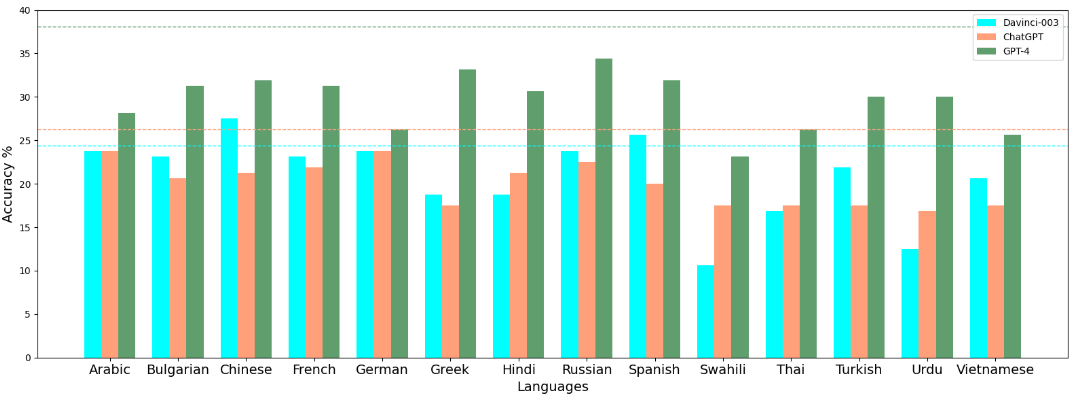}
  \caption{We illustrate the session-based results of closed-source LLMs in the creating new slides task, where the instructions are translated into 14 non-English languages. The bar for each language represents the LLM's accuracy in the corresponding language setting. The dotted line is the LLM's accuracy when tested in the English setting. }
  \label{fig:sess_task1}
\vspace{-10pt}
\end{figure*}

\begin{figure*}[ht]
  \centering
  \includegraphics[width=1\textwidth]{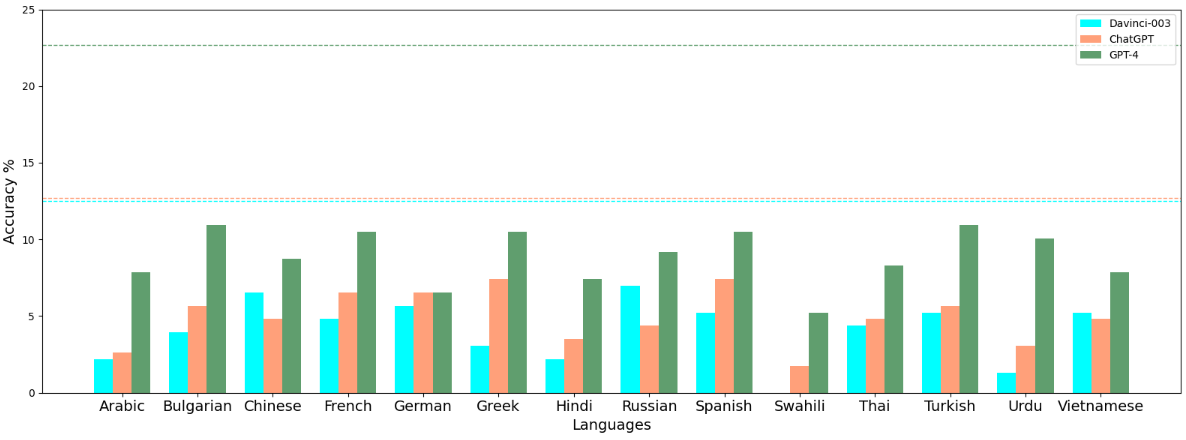}
  \caption{We illustrate the turn-based results of closed-source LLMs in the editing template task, where the instructions are translated into 14 non-English languages. The bar for each language represents the LLM's accuracy in the corresponding language setting. The dotted line is the LLM's accuracy when tested in the English setting.}
  \label{fig:turn_task2}
\vspace{-10pt}
\end{figure*}

We report the session-based performance of the creating new slides task in Figure~\ref{fig:sess_task1}.
For the editing template task, we report the turn-based accuracy of 3 LLMs for it in Figure~\ref{fig:turn_task2}. We find that all LLM's session-based accuracy in this task is smaller than 4 percent. So we do not further report and analyze the session-based result.
\section{Detailed Wrong Examples Made by LLMs}
\label{appendix,wrong_example}
We provide 4 typical wrong examples with their explanations in Figure~\ref{fig:wrong_examples}.
\begin{figure*}[!t]
\centering
\tcbset{colback=white!15!white, colframe=orange!85!red, fonttitle=\bfseries}
\begin{tcolorbox}[title=Calling unavailable APIs or new APIs \qquad\qquad Misunderstanding the instruction,sidebyside, sidebyside align=top, coltitle=black,]
\textbf{Example 1 (GPT-4 in the API update setting):}\\\textbf{Instruction}: Fill the star with blue color.
\\\\
\textbf{Generated API sequence}: \\
choose$\_$shape('star')
set$\_$shape$\_$gradient('star', 'radial', ['blue', 'light blue'])
\\\\
\textbf{Explanation:} We do not provide the API set$\_$shape$\_$gradient in the API reference file.
\\\\
\textbf{Example 2 (ChatGPT in the API update setting)}"\\\textbf{Instruction}: Insert one clock. Make it on the right side. 
\\\\
\textbf{Generated API sequence}: \\
insert$\_$date$\_$time("h:mm AM/PM", "right")\\\\
\textbf{Label API sequence}: \\
insert$\_$picture("clock")\\
align$\_$slide$\_$right()
\\\\\textbf{Explanation}: The new API insert$\_$date$\_$time provided in the reference file is related to the clock (time). But it's not the right API. 
\\\\
\tcblower
\textbf{Example 1 (ChatGPT in the sentence-level robustness setting)}
\\\textbf{Instruction}: You know, I've always admired the creativity and innovation behind successful business plans. Anyway, go ahead and create a slide and insert the title "Business Plan". By the way, have you ever noticed how soothing the color blue can be? It's one of my favorites.
\\\\
\textbf{Generated API sequence}: \\
create$\_$slide()\\
choose$\_$title()\\
insert$\_$text("Business Plan")\\
set$\_$background$\_$color("blue")
\\\\\textbf{Explanation}: The LLM misunderstands the instruction and generates extra 'set$\_$background$\_$color("blue")' API action, which leads to the wrong prediction.\\\\
\\\textbf{Example 2 (GPT-4 in the sentence-level robustness setting)}"\\\textbf{Instruction}: I hope you're having a great day! By the way, the height of the picture should be 2 on slide 6, and it's incredible how technology can help us with these tasks. While you're at it, the font size of all textboxes should be 10 on slide 11. Time flies when you're working on presentations, doesn't it?
\\\\
\textbf{Generated API sequence}: \\
move$\_$to$\_$slide(6)\\
choose$\_$picture(0)\\
set$\_$height(2)\\
\\\\\textbf{Explanation}: The LLM misunderstands the instruction and forgets to set the font size of all textboxes to be 10 on slide 11.
\\\\
\end{tcolorbox}
\captionof{figure}{The detailed wrong examples with their explanations.}
\label{fig:wrong_examples}
\end{figure*}
\end{document}